\documentclass{article} 
\usepackage{iclr2026_conference,times}

\usepackage{amsmath,amsfonts,bm}

\def\eqref#1{equation~\ref{#1}}

\def\1{\bm{1}}

\DeclareMathAlphabet{\mathsfit}{\encodingdefault}{\sfdefault}{m}{sl}
\SetMathAlphabet{\mathsfit}{bold}{\encodingdefault}{\sfdefault}{bx}{n}

\definecolor{softblue}{rgb}{0.21,0.49,0.74}

\usepackage[breaklinks,colorlinks,allcolors=softblue]{hyperref}
\usepackage{url}
\usepackage{booktabs}       
\usepackage{amsfonts}       
\usepackage{nicefrac}
\usepackage{microtype}

\usepackage{algorithm}
\usepackage{algorithmic}
\usepackage{amsmath}
\usepackage{amssymb}
\usepackage{mathtools}
\usepackage{bm}
\usepackage{multirow}
\usepackage{multicol}

\usepackage{colortbl}
\usepackage{makecell}

\newcommand{\ie}{\textit{i}.\textit{e}.}

\usepackage{newfloat}
\usepackage{listings}

\usepackage[utf8]{inputenc}
\usepackage[T1]{fontenc}
\usepackage{newtxtext}

\usepackage{courier} 

\title{\textit{ImagerySearch}: Adaptive Test-Time Search for Video Generation Beyond Semantic Dependency Constraints}

\author{%
    \textbf{Meiqi Wu$^{1,3}$\thanks{Work done during the internship at AMAP, Alibaba Group.} \quad}
    \textbf{Jiashu Zhu$^{2}$\quad}
    \textbf{Xiaokun Feng$^{1,3}$ \quad} 
    \textbf{Chubin Chen$^{4}$ \quad} 
    \textbf{Chen Zhu$^{5}$ \quad}\\
    \textbf{Bingze Song$^{2}$\quad}
    \textbf{Fangyuan Mao$^{2}$\quad}
    \textbf{Jiahong Wu$^{2}$\thanks{Project leader.} \quad}
    \textbf{Xiangxiang Chu$^{2}$\quad} 
    \textbf{Kaiqi Huang$^{1,3}$\thanks{Corresponding author.}}\\
    $^1$UCAS \ 
    $^2$AMAP, Alibaba Group \ 
    $^3$CRISE \ 
    $^4$THU \ 
    $^5$SEU  \ 
    \vspace{0.5em} \\ 
    \textbf{GitHub:}\ \href{https://github.com/AMAP-ML/ImagerySearch/}{%
        \color{magenta}\selectfont 
        \texttt{https://github.com/AMAP-ML/ImagerySearch/}%
    }
    \vspace{-1em} 
}

\begin{document}

\maketitle

\begin{figure}[!ht]
    \centering
    \includegraphics[width=\linewidth]{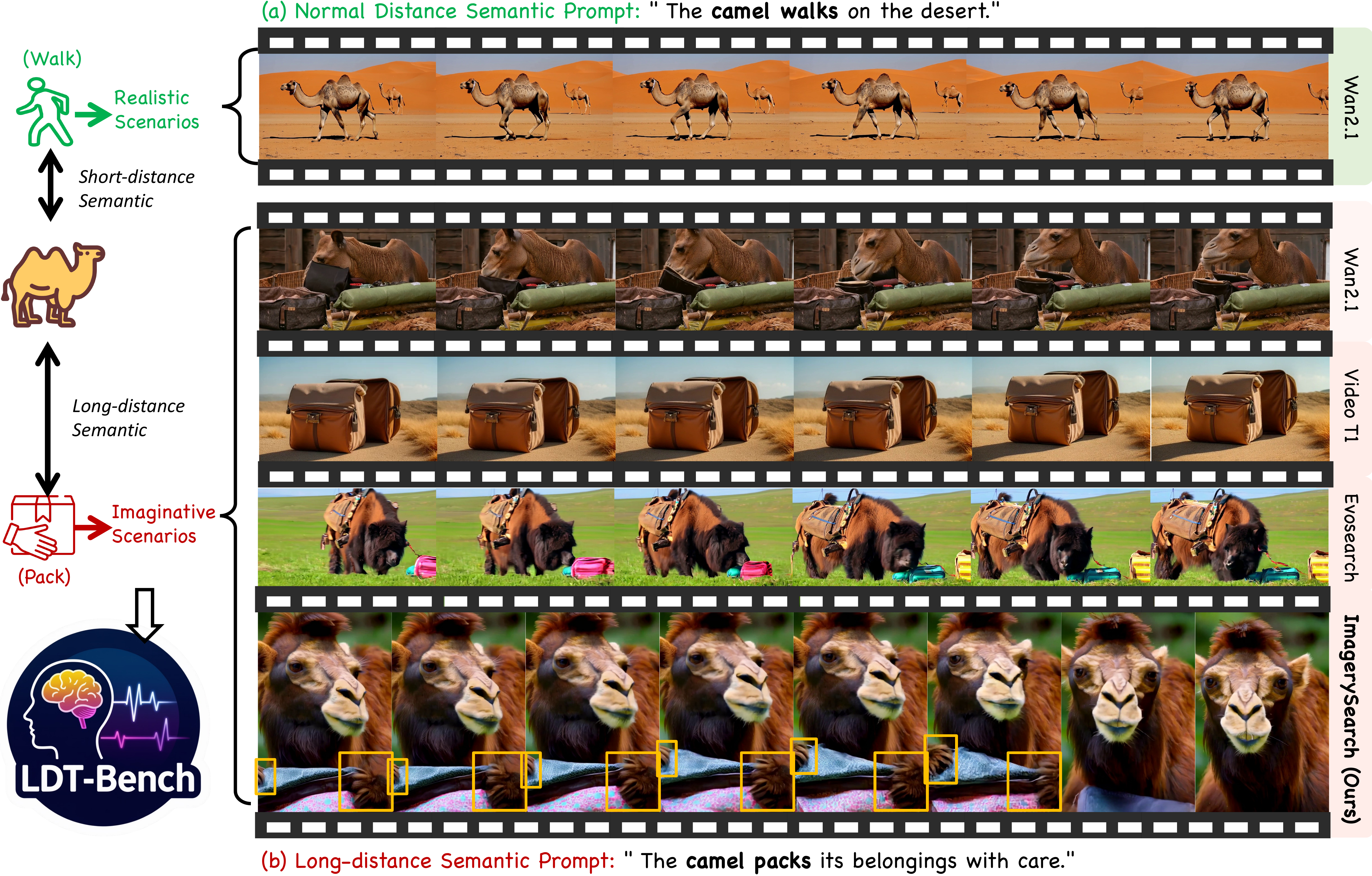}
    \caption{\textbf{The motivation of ImagerySearch.} The figure illustrates two semantic dependency scenarios related to camels. \textbf{Left}: The distance depicts the corresponding strength of prompt tokens during the denoising process. \textit{LDT-Bench} consists of imaginative scenarios with long-distance semantics, whose semantic dependencies are typically weak. \textbf{Right}: Wan2.1 performs well on short-distance semantics but fails under long-distance. Test time scaling methods ($e.g.$, Video T1~\citep{liu2025videot1}, Evosearch~\citep{he2025evosearch}) also struggle. However, \textit{ImagerySearch} generates coherent, context-aware motions (orange box).}
    \label{fig:intro}
\end{figure}

\begin{abstract}
Video generation models have achieved remarkable progress, particularly excelling in realistic scenarios; however, their performance degrades notably in imaginative scenarios. These prompts often involve rarely co-occurring concepts with long-distance semantic relationships, falling outside training distributions. Existing methods typically apply test-time scaling for improving video quality, but their fixed search spaces and static reward designs limit adaptability to imaginative scenarios.
To fill this gap, we propose \textbf{ImagerySearch}, a prompt-guided adaptive test-time search strategy that dynamically adjusts both the inference search space and reward function according to semantic relationships in the prompt. This enables more coherent and visually plausible videos in challenging imaginative settings. To evaluate progress in this direction, we introduce \textbf{LDT-Bench}, the first dedicated benchmark for long-distance semantic prompts, consisting of 2,839 diverse concept pairs and an automated protocol for assessing creative generation capabilities.
Extensive experiments show that ImagerySearch consistently outperforms strong video generation baselines and existing test-time scaling approaches on LDT-Bench, and achieves competitive improvements on VBench, demonstrating its effectiveness across diverse prompt types. We will release LDT-Bench and code to facilitate future research on imaginative video generation.
\end{abstract}

\vspace{-3mm}
\section{Introduction}
\label{sec:intro}

Imagine describing a surreal scene--``a panda playing violin on Mars during a sandstorm''--and instantly seeing it come to life as a video. Text-to-video generation promises just that: the ability to turn language into vivid, dynamic worlds. Recent video generation models have made significant progress in generating realistic scenes~\citep{wang2023modelscope, yang2024cogvideox, opensora2, Sora, wan2.1}; however, their performance drops sharply when handling subjectively imaginative scenarios, hindering the advancement of truly creative video generation. \textit{Why is imagination so hard to generate?}

This limitation arises from two primary factors. \textbf{(1) The model's semantic dependency}: Generative models exhibit strong semantic dependency constraints on long-distance semantic prompts, making it difficult to generalize to imaginative scenarios beyond the training distribution (Fig.~\ref{fig:intro}). \textbf{(2) The scarcity of imaginative training data}: Mainstream video datasets~\citep{huang2024vbench,liu2024evalcrafter,sun2024t2vcompbench,liu2023fetv,liao2025devil,ling2025vmbench} predominantly contain realistic scenarios, offering limited imaginative combinations characterized by long-distance semantic relationships (Fig.~\ref{fig:ldt-bench}(d)). Recent test-time scaling approaches~\citep{liu2025videot1, he2025evosearch} alleviate data scarcity by sampling multiple candidates and selecting the most promising one. However, their predefined sampling spaces and static reward functions constrain adaptability to the open-ended nature of creative generation.

The Imagery Construction theory~\citep{thomas1999theories,pylyshyn2002mental} posits that humans create mental scenes for imaginative scenarios by iteratively refining visual imagery in response to language. Motivated by this principle, we introduce \textbf{ImagerySearch}, a test-time search strategy that enhances prompt-based visual generation. ImagerySearch comprises two core components: (i) \textbf{S}emantic-distance-\textbf{a}ware \textbf{D}ynamic \textbf{S}earch \textbf{S}pace (\textbf{SaDSS}), which adaptively modulates sampling granularity according to the semantic span of the prompt; and (ii) \textbf{A}daptive \textbf{I}magery \textbf{R}eward (AIR), which incentivizes outputs that align more closely with the intended semantics. 

To assess generative models in imaginative settings, we propose \textbf{LDT-Bench}, the first benchmark designed specifically for long-distance semantic prompts. It comprises 2,839 challenging concept pairs, constructed by maximizing semantic distance across object–action and action–action dimensions from diverse recognition datasets ($e.g.$, ImageNet-1K~\citep{deng2009imagenet}, Kinects-600~\citep{jo2018Kinects600}). In addition, LDT-Bench includes an automatic evaluation protocol, \textbf{ImageryQA}, which quantifies creative generation with respect to element coverage, semantic alignment, and anomaly detection.

Extensive experiments reveal that general models ($e.g.$, Wan14B~\citep{wan2.1}, Hunyuan-13B~\citep{kong2024hunyuanvideo}, CogVideoX~\citep{yang2024cogvideox}) and TTS-based models ($e.g.$, VideoT1~\citep{liu2025videot1}, EvoSearch~\citep{he2025evosearch}) suffer from significant degradation in video quality and semantic alignment when conditioned on long-distance semantics. In contrast, our framework consistently improves generation fidelity and alignment, demonstrating superior capability in handling long-distance semantic prompts. 

Our contributions can be summarized as follows:
\begin{itemize}
    \item We propose ImagerySearch, a dynamic test-time scaling law strategy inspired by mental imagery that adaptively adjusts the inference search space and reward according to prompt semantics.
    
    \item We present LDT-Bench, the first benchmark specifically designed for video generation from long-distance semantic prompts. It comprises 2,839 prompts--spanning 1,938 subjects and 901 actions--and offers an automatic evaluation framework for assessing model creativity in imaginative scenarios.
  
    \item Extensive experiments on LDT-Bench and VBench reveal that our approach consistently improves imaging quality and semantic alignment under long-distance semantic prompts.
    
\end{itemize}

\section{Related Work}
\label{sec:rework}
\noindent\textbf{Text-to-Video Generation Models.} With advances in generative modeling ~\citep{ho2020denoising,chu2024visionllama,lei2025advancingendtoendpixelspace, chu2025usp, chen2025s} and increased training resources, large-scale T2V models~\citep{Sora, kling, Gen3, Vidu, opensora, opensora2, genmo2024mochi, kong2024hunyuanvideo, wan2.1} have emerged, capable of generating coherent videos, understanding physics, and generalizing to complex scenarios. But they require massive data, and collecting enough long-range semantic prompts is impractical.  
Although fine-tuning~\citep{fan2023optimizing, lee2023aligning, black2023training, wallace2024diffusion, clark2023directly, domingo2024adjoint, mao2025omni} and post-training~\citep{yuan2024instructvideo, prabhudesai2024video, luo2023latent, li2024t2v, li2024t2v2} methods mitigate data requirements to some extent, the extreme scarcity of long-distance semantic videos still hinders effective training. In contrast, the Test-Time Scaling (TTS) methods~\citep{oshima2025inference, xie2025sana, yang2025scalingnoise, liu2025videot1, he2025evosearch} used in ImagerySearch require no additional training and achieve strong performance through a highly general approach.

\noindent\textbf{Test-Time Scaling in T2V Models.} TTS improves performance by using rewards to select better outputs~\citep{jaech2024openai, guo2025deepseek}. In T2V generation, TTS are primarily explored in two aspects: selection strategies and reward strategies.
Selection strategies mainly include Best-of-N, particle sampling, and beam search. The Best-of-N~\citep{ma2025inference,liu2025videot1} selects the top N outputs from multiple generations. Particle sampling~\citep{fk,li2024derivative,Li2025DynamicSF,Singh2025CoDeBC,kim2025testtime} improves upon this by performing importance-based sampling across the denoising process. Beam search \citep{liu2025videot1, yang2025scalingnoise,xie2025sana,oshima2025inference,liu2025videot1, he2025scaling} keeps multiple candidates at each step, expanding the sequence set over time. Reward strategies are based on various evaluation metrics, such as VisionReward~\citep{xu2024visionreward}, ImageReward~\citep{xu2023imagereward}, Aesthetic score~\citep{schuhmann2022laion}, which guide the selection process by quantifying the quality of generated output. These reward functions are crucial for aligning outputs with desired visual and semantic characteristics.

Current TTS methods optimize search and reward strategies for general T2V generation to enhance overall performance. In this work, we investigate this specific challenge and explore how TTS can be leveraged to improve model performance in long-distance semantic prompts.

\noindent\textbf{Evaluation of Video Generative Models.} Early video-generation metrics are simplistic: some diverged from human judgment \citep{unterthiner2018towards,radford2021learning}, while others reused real-video tests unsuited to synthetic clips \citep{soomro2012ucf101,xu2016msr}.
Later, studies~\citep{szeto2022devil, liu2023fetv, liu2024evalcrafter, huang2024vbench++, sun2025t2v, zheng2025vbench, chen2025finger} such as VBench~\citep{huang2024vbench} evaluated AI-generated videos from a comprehensive, multi-dimensional perspective.
Several studies~\citep{liu2024fr,yuan2024chronomagic,yuan2025opens2v,ling2025vmbench} refine evaluation along single dimensions such as frame realism or temporal coherence.

Although existing methods focus on video quality and human perception, semantic content assessment remains underexplored. Current benchmarks struggle to effectively evaluate long-distance semantic prompts, which are key to advancing video generation capabilities. To address this, LDT-Bench was introduced as the first benchmark for evaluating long-distance semantic understanding in video generation.

\section{ImagerySearch}
\label{sec:method}

\begin{figure*}
    \centering
    \includegraphics[width=\linewidth]{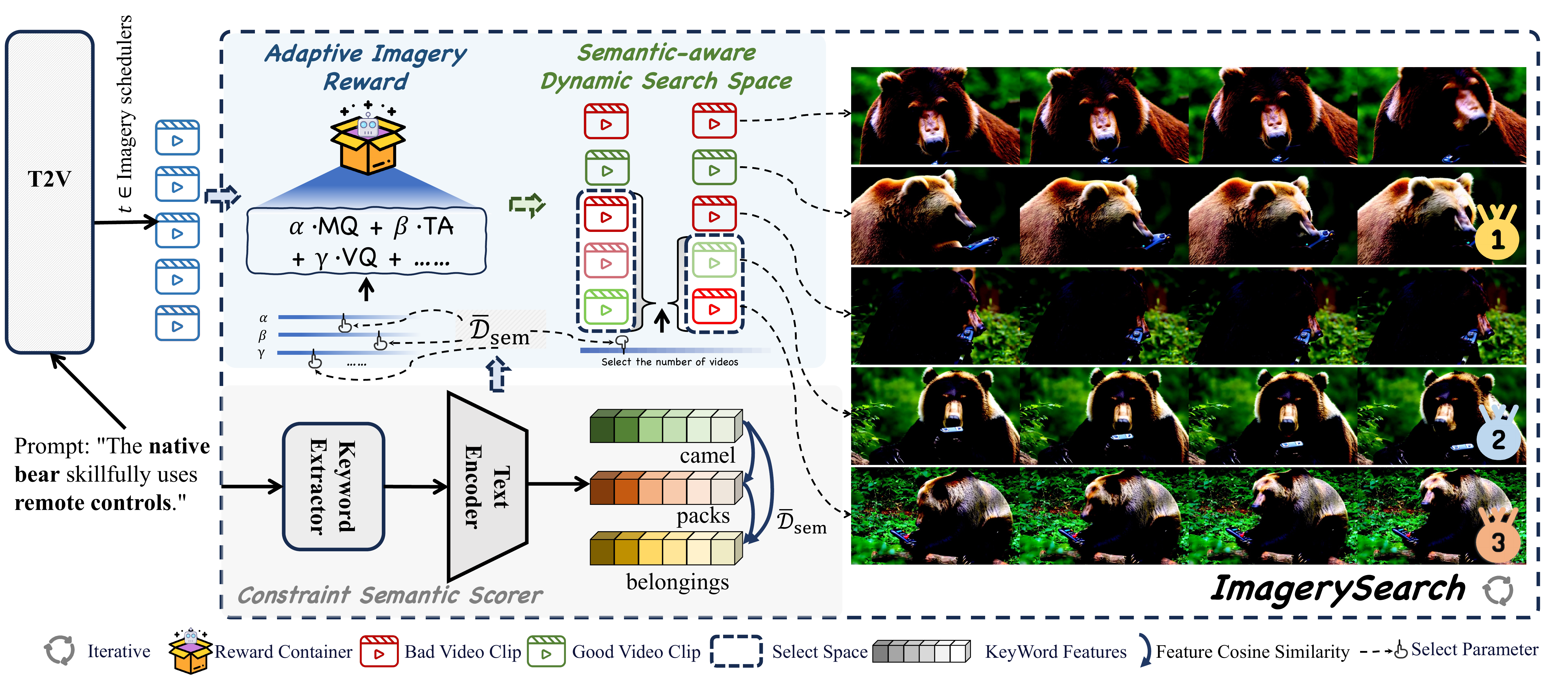}
    \vspace{-3mm}
    \caption{Overview of our ImagerySearch. The prompt is scored by the Constrained Semantic Scorer (producing \(\bar{\mathcal{D}}_{\text{sem}}\)) and simultaneously fed to the T2V backbone (Wan2.1). At every step \(t\) specified by the imagery scheduler, we sample a set of candidate clips, rank them with a reward function conditioned on \(\bar{\mathcal{D}}_{\text{sem}}\), and retain only a \(\bar{\mathcal{D}}_{\text{sem}}\)-controlled subset. The loop repeats until generation completes.}
    \vspace{-3mm}
    \label{fig:method}
\end{figure*}

Text-to-video generation aims to synthesize coherent videos conditioned on prompts. Diffusion models inherently possess the flexibility to adjust test-time computation via the number of denoising steps. To further improve generation quality, we formulate a search problem that identifies better noise inputs for the diffusion sampling process. We organize the design space along two axes: the reward functions that evaluate video quality, and the search algorithms that explore and select optimal noise candidates.

\subsection{Preliminaries}
In standard diffusion frameworks, sampling starts from Gaussian noise $\mathbf{x}_{T} \sim \mathcal{N}(0, \mathbf{I})$, and the model iteratively denoises the latent through a learned network $f_{\theta}$.
As a widely used sampling paradigm, DDIM performs the following step-wise denoising update:

\begin{equation}
\mathbf{x}_{t-1} = \zeta_{t-1}( \frac{ \mathbf{x}_{t}- \sigma_{t} f_{\theta}(\mathbf{x}_t, t, \mathbf{c})}{\zeta_{t}} ) + \sigma_{t-1} f_{\theta}(\mathbf{x}_t, t, \mathbf{c}),
\end{equation}
Where $\zeta_{t-1}$, $\zeta_{t}$, $\sigma_{t-1}$ denote predefined schedules.

Prior test-time scaling approaches \citep{liu2025videot1, he2025evosearch, yang2025scalingnoise} operate within a fixed noise search space and use static reward functions--such as VideoScore \citep{he2024videoscore}, VideoAlign \citep{liu2025videoalign}, or their combinations--to rank candidates. By contrast, our framework supports flexible reward design and adaptive noise selection, substantially improving both sample efficiency and generation quality.

\subsection{Dynamic Search Space}

Inspired by imagery cognitive theory~\citep{thomas1999theories,pylyshyn2002mental,feng2023hierarchical}—which posits that humans expend more effort and time to construct mental imagery for semantically distant concepts—we likewise adapt the candidate-video search space to a prompt’s semantic distance: shrinking it for short-distance prompts to boost test-time efficiency, and enlarging it for long-distance prompts to explore a broader range of possibilities. Therefore, we propose a \textbf{S}emantic-distance-\textbf{a}ware \textbf{D}ynamic \textbf{S}earch \textbf{S}pace (SaDSS). 

As shown in Fig.~\ref{fig:method}, this adaptive resizing is driven by a \textbf{Constrained Semantic Scorer}, which dynamically modulates the search space. Specifically, we define semantic distance as the average embedding distance between key entities (objects and actions) extracted from the prompt. Given a prompt $\mathbf{p}$, we extract its compositional units $\{p_i\}_{i=1}^{n}$ and compute:
\begin{equation}
\bar{\mathcal{D}}_{\text{sem}}(\mathbf{p}) = \frac{1}{|E|} \sum_{(i,j)\in E} \left\| \phi(p_i) - \phi(p_j) \right\|_2,
\end{equation}
where $\phi(\cdot)$ denotes the embedding function ($e.g.$, T5 encoder), and $E$ is the set of key entity pairs in the prompt.

At inference time, we adapt the sampling procedure based on $\bar{\mathcal{D}}_{\text{sem}}$. Specifically, the search space dynamically adapts based on semantic distance. Formally, the number of candidates $N_t$ at timestep $t$ is dynamically adjusted as:
\begin{equation}
N_t = N_{\text{base}} \cdot \left(1 + \lambda \cdot \bar{\mathcal{D}}_{\text{sem}}(\mathbf{p})\right),
\end{equation}
where $N_{\text{base}}$ is the base number of samples, and $\lambda$ is a scaling factor that controls the sensitivity to semantic distance. In this work, we set $\lambda = 1$.

By tailoring the search scope to the inherent difficulty of the prompt, SaDSS encourages the model to explore more diverse visual hypotheses when needed, improving visual plausibility under challenging conditions, without incurring unnecessary computational costs for simple prompts.
 
\subsection{Adaptive Imagery Reward}

Based on our observations, adjacent denoising steps alter the latent video only marginally, so we invoke ImagerySearch at a few key noise levels $\mathcal S=\{5,\,10,\,20,\,45\}$, termed the \emph{Imagery Schedule} (see Appendix A). As shown in Fig.~\ref{fig:method}, starting from a partially denoised latent $\mathbf{x}_t$, we produce $\hat{\mathbf{x}}_0$ by completing the denoising trajectory and compute the reward on $\hat{\mathbf{x}}_0$ to assess the influence of different denoising stages on the final video quality.

To enhance semantic alignment between generated videos and prompts with long-distance semantics, we introduce an \textbf{A}daptive \textbf{I}magery \textbf{R}eward (AIR) that modulates evaluation feedback based on the prompt’s semantic difficulty. Specifically, we incorporate the semantic distance as a soft re-weighting factor into the reward formulation. The reward $R_{\text{AIR}}(\hat{\mathbf x}_{0})$ for each candidate video $\mathbf{x}_{0}$ is defined as:
\begin{equation}
R_{\text{AIR}}(\hat{\mathbf x}_{0}) = (\alpha \cdot \mathrm{MQ} +\beta \cdot \mathrm{TA} + \ \gamma \cdot \mathrm{VQ} + \omega \cdot R_{any})\cdot\bar{\mathcal{D}}_{\text{sem}}(\hat{\mathbf x}_{0}),
\end{equation}
where $\alpha$, $\beta$, $\gamma$, and $\omega$ are scaling factors that adaptively adjust the reward based on the prompt semantic distance $\bar{\mathcal{D}}_{\text{sem}}$. $\mathrm{MQ}$, $\mathrm{TA}$, and $\mathrm{VQ}$ are from VideoAlign~\citep{liu2025videoalign}, and $R_\text{any}$ denotes an extensible reward ($e.g.$, VideoScore~\citep{he2024videoscore}, VMBench~\citep{ling2025vmbench}, UnifiedReward~\citep{wang2025unified}, VisionReward~\citep{xu2024visionreward}).

\section{LDT-Bench}
\label{sec:ldt}

\begin{figure*}[t!]
    \centering
    \includegraphics[width=0.9\linewidth]{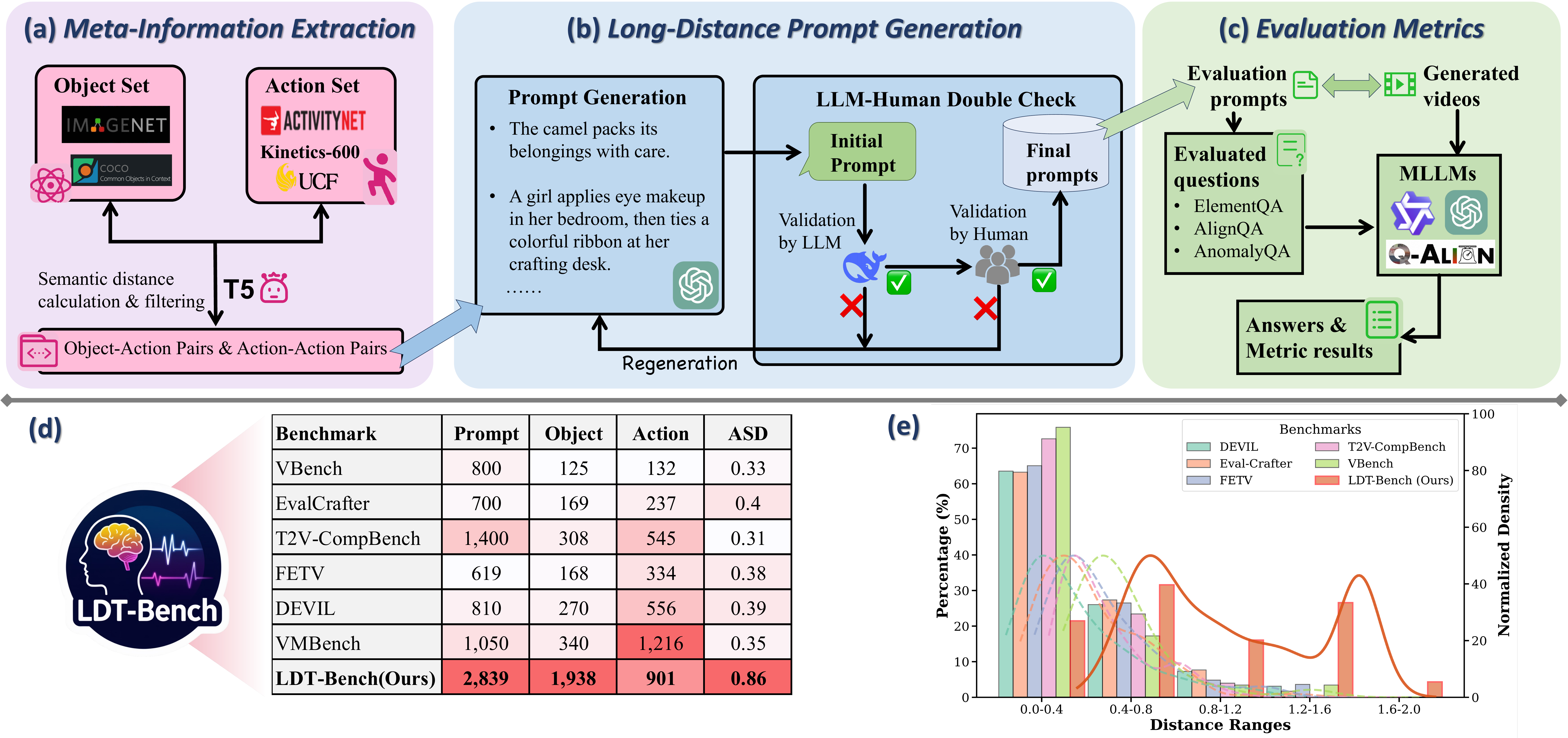}
    \vspace{-3mm}
    \caption{Overview of our LDT-Bench. \textbf{Upper}: (a) LDT-Bench is built by first extracting meta-information from existing recognition datasets; (b) GPT-4o is then used to generate candidate prompts, which are filtered jointly by DeepSeek and humans to obtain the final prompt set; (c) Additionally, we design a set of three MLLM-based QA tasks that serve as the creativity metric. \textbf{Lower}: (d) Compared with other benchmarks, LDT-Bench covers a much richer variety of categories; (e) its prompts also exhibit a semantic-distance distribution that is shifted toward substantially longer ranges. Note that ``ASD'' denotes the average semantic distance of prompts.}
    \vspace{-3mm}
    \label{fig:ldt-bench}
\end{figure*}

The rapid progress of video generation models is closely tied to the development of targeted evaluation benchmarks. Existing benchmarks primarily assess models using text prompts designed to depict realistic scenarios. 
However, as video generation models have achieved impressive performance in realistic scenarios, it is timely to shift the focus towards imaginative scenarios. Generally, such complex settings involve prompts in which entities--such as objects and actions--exhibit long semantic distances, meaning these entities rarely co-occur ($e.g.$, ``a panda piloting a helicopter''). These corner cases reveal the robustness limits of generative models. Nonetheless, most existing works remain limited to qualitative analysis on a few cases, and there is a lack of a unified benchmark specifically designed for this task.

To fill this gap, we propose a novel benchmark \textbf{LDT-Bench}, designed to systematically analyze the generalization ability of video generation models in complex scenarios induced by prompts with \textbf{L}ong-\textbf{D}istance semantic \textbf{T}exts.
In the following sections, LDT-Bench is introduced from two perspectives: the construction of the prompt suite and the design of evaluation metrics.
The core components of LDT-Bench are illustrated in Fig.~\ref{fig:ldt-bench}.

\subsection{Prompt Suite}

\noindent\textbf{Meta-information Extraction.} Considering that objects and actions are the main entities in text prompts, we construct our prompts using the following two structural types. (1) \textbf{Object--Action}: An object combined with an uncommon or incompatible action. (2) \textbf{Action--Action}: Two semantically distant or even contradictory actions.

To cover a wide range of objects and actions, we build our object and action sets from representative large-scale datasets. Specifically, the object set is derived from ImageNet-1K \citep{deng2009imagenet} and COCO \citep{lin2014coco} (covering 1,938 objects), while the action set is collected from ActivityNet \citep{caba2015activitynet}, UCF101 \citep{soomro2012ucf101}, and Kinetics-600 \citep{jo2018Kinects600} (covering 901 actions). These collections serve as the foundation for subsequent prompt generation.

We first encode each object and action element $\mathrm{text}_i$ using a pretrained T5 text encoder~\citep{raffel2020exploring}, obtaining a high-dimensional textual feature $\mathbf{h}_i \in \mathbb{R}^d$. These embeddings are then projected into a shared 2D semantic space via Principal Component Analysis (PCA):
\begin{equation}
\mathbf{z}_i = \mathrm{PCA}(\mathbf{h}_i) = \mathrm{PCA}(\mathrm{T5}(\mathrm{text}_i)), \quad \mathbf{z}_i \in \mathbb{R}^2,
\end{equation}
where $\mathbf{z}_i$ represents the semantic position of the $i$-th element in the 2D space. T5 can be replaced with other encoders, such as CLIP~\citep{radford2021learning}; see Appendix B.1 for details.

To measure semantic divergence, we compute the Euclidean distance between each pair of elements as a criterion for selecting long-distance semantic prompts.
We then construct two candidate sets: one by pairing each object with its most distant action (1,938 object--action pairs), and the other by matching each action with its most distant counterpart (901 action--action pairs). From each set, we select the 160 most distant pairs, resulting in 320 high-distance prompts that challenge the model with long-distance semantic combinations. For more analysis of the prompt suite, please refer to Appendix B.2.

\noindent\textbf{Long-distance Prompt Generation.} Based on the obtained text element pairs, we employ a large language model, \ie,  GPT-4o~\citep{hurst2024gpt}, to generate fluent and complete text prompts by filling in necessary sentence components. Subsequently, each prompt is double-checked by both DeepSeekR1~\citep{guo2025deepseek} and human annotators to ensure quality, resulting in our final prompt suite. 
The detailed generation process and several illustrative cases are presented in Fig.~\ref{fig:ldt-bench} (b).

\subsection{Imagery Evaluation Metrics}
To quantitatively evaluate the performance of video generation models under long-distance semantic settings, we develop targeted evaluation metrics. Inspired by recent MLLMs-based evaluation methods \citep{cho2023davidsonian,feng2025narrlv}, we generate questions based on the text prompts. 
Subsequently, MLLMs with strong semantic comprehension capabilities analyze the generated videos in response to these questions, yielding quantitative evaluation results.
Specifically, our assessment framework encompasses three primary dimensions.

\noindent\textbf{ElementQA.} Because our prompts focus on objects and actions, ElementQA primarily consists of targeted questions revolving around these elements. 
For example, given the prompt ``The traffic light is dancing.'', we can generate two questions: ``Does the traffic light appear in the video?'' and ``Is the traffic light performing a dancing action?''

\noindent\textbf{AlignQA.} In addition to the basic semantic information covered by ElementQA, we also evaluate the generated videos in terms of visual quality and aesthetics \citep{murray2012ava}.
Given the challenging and inherently subjective nature of this assessment, we employ recently developed MLLMs that have been specifically optimized for alignment with human perception to perform the evaluation \citep{huang2024aesexpert,wu2023q}.

\noindent\textbf{AnomalyQA.} We have observed that current video generation models frequently produce anomalous outputs. Consequently, we also leverage MLLMs to analyze the generated frames and answer targeted questions aimed at identifying these anomalies.

\noindent\textbf{Implementation Details.} For ElementQA, we employ Qwen2.5-VL-72B-Instruct~\citep{qwen2.5} as the underlying MLLM, whereas for AlignQA we adopt Q-Align \citep{wu2023q}, a model specifically optimized for rating visual quality and aesthetics. Given the broader generalization required by AnomalyQA, we utilize the more powerful GPT-4o \citep{openai2024gpt4o} for evaluation. We collectively refer to these three components as ImageryQA. Further implementation details are provided in Appendix B.3.

\section{Experiments}
\label{sec:exper}

\subsection{Experimental Setup}

\noindent\textbf{Datasets \& Metrics.} To assess the imaginative capacity of video-generation models, we evaluate them on both LDT-Bench and VBench~\citep{huang2024vbench}, using each benchmark’s full prompt suite and associated metrics. 

\noindent\textbf{Compared Models.} We compare two categories of models: (1) \textit{General models}: Hunyuan~\citep{kong2024hunyuanvideo}, Wan2.1~\citep{wan2.1}, Open-Sora~\citep{zheng2024opensora}, CogVideoX~\citep{yang2024cogvideox}; (2) \textit{TTS methods}: Video-T1~\citep{liu2025videot1} and EvoSearch~\citep{he2025evosearch}. We use Wan2.1 as the base model and generate 33-frame clips with the default settings (see Appendix C for details). 

\noindent\textbf{Experimental Environment.} All experiments are run on a server equipped with 8 $\times$ NVIDIA H20 GPUs (96 GB each), an Intel Xeon Gold 6348 CPU (32 cores, 2.6 GHz), and 512 GB of RAM, under Ubuntu 20.04 LTS (kernel 5.15). We used Python 3.9 with PyTorch 2.5.1 (CUDA 12.4, cuDNN 9.1), torchvision 0.20.1, and Transformers 4.50.3.

\subsection{Comparison with Other Generation Models}

\noindent\textbf{Performance on LDT-Bench.} As shown in Tab.~\ref{tb:LDT-bench}, we adopt Wan2.1 as the base model. Our method achieves a significant improvement of 8.83\%, demonstrating a clear advantage. Furthermore, compared to other test-time scaling approaches, ImagerySearch also delivers consistently superior performance. These results highlight the effectiveness of our method in handling long-distance semantic prompts and its robustness in imagination-driven scenarios.

\begin{figure*}[t!]
    \centering
    \includegraphics[width=0.9\linewidth]{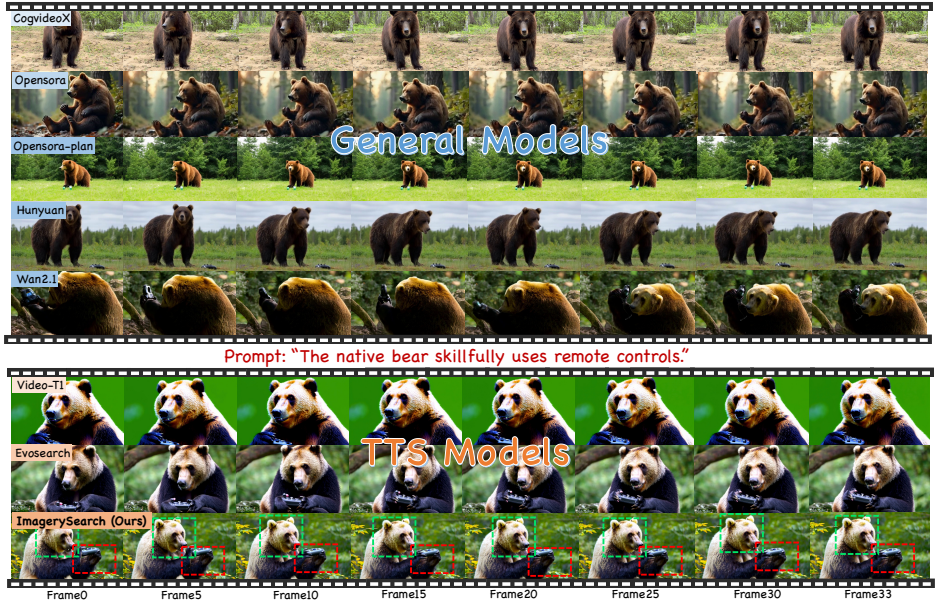}
    \caption{Visualization of examples. \textbf{Upper}: Results from general models. \textbf{Lower}: ImagerySearch versus other test-time scaling methods. Ours produces more vivid actions under long-distance semantic prompts.}
    \label{fig:visual_examples}
\end{figure*}

\begin{table}[!ht]
\centering
\resizebox{0.7\linewidth}{!}{
\begin{tabular}{l|cccc}

\toprule
        & \multicolumn{4}{c}{LDT-Bench (\%) $\uparrow$} \\ \cmidrule{2-5}
\multicolumn{1}{c|}{\multirow{-2}{*}{Model}}
        & ElementQA & AlignQA & AnomalyQA & ImageryQA (All) \\ \midrule
Wan2.1~\citep{wan2.1}
        & 1.66
        & 31.62
        & 15.00
        & 48.28 \\ \midrule
Video-T1~\citep{liu2025videot1}
        & 1.91
        & \textbf{38.16}
        & 14.68
        & 54.75 \\
Evosearch~\citep{he2025evosearch}
        & 1.92
        & 36.10
        & 16.46
        & 54.48 \\ \midrule
\textbf{ImagerySearch (Ours)}
        & \textbf{2.01}
        & 36.82
        & \textbf{18.28}
        & \textbf{57.11} \\ \bottomrule
\end{tabular}
}

\caption{Quantitative comparison on LDT-Bench. ImagerySearch achieves the best average performance.}

\label{tb:LDT-bench}
\end{table}

\begin{table*}[!htb]
\centering
\resizebox{\linewidth}{!}{
\begin{tabular}{cl|cccccc|c}
\toprule
\multicolumn{2}{c|}{} & \multicolumn{7}{c}{VBench (\%) $\uparrow$}
\\ \cmidrule{3-9} 
\multicolumn{2}{c|}{\multirow{-2}{*}{Model}} & \makecell[c]{Aesthetic\\Quality}    & \makecell[c]{Background\\Consistency}   & \makecell[c]{Dynamic\\Degree}          & \makecell[c]{Imaging\\Quality}     & \makecell[c]{Motion\\Smoothness}    & \makecell[c]{Subject\\Consistency}  & Average   \\ \midrule
                          & Wan2.1~\citep{wan2.1}                        & 50.50          & 91.80          & 82.85          & 58.25          & 97.50          & 90.25          & 78.53          \\
                          & Opensora~\citep{opensora2}                      & 48.80          & 95.25          & 73.15          & 61.35          & 99.05          & 92.95          & 78.43          \\
                          & CogvideoX~\citep{yang2024cogvideox}                     & 48.80          & 95.30          & 47.20          & 65.05          & 98.55          & 94.65          & 74.93          \\
\multirow{-5}{*}{General} & Hunyuan~\citep{kong2024hunyuanvideo}                       & 50.45          & 92.65          & 85.00          & 59.55          & 95.75          & 90.55          & 78.99          \\ \midrule
                          & Video-T1~\citep{liu2025videot1}                      & 57.20          & 95.65          & 54.05          & 60.25          & \textbf{99.30} & 94.80          & 76.88          \\
                          & Evosearch~\citep{he2025evosearch}                     & 55.55          & 94.80          & 80.95          & 68.90          & 97.70          & 94.55          & 82.08          \\ \cmidrule{2-9} 
\multirow{-3}{*}{TTS}     & \textbf{ImagerySearch (Ours)} & \textbf{57.70} & \textbf{96.00}          & \textbf{84.05} & \textbf{69.20} & 98.00          & \textbf{95.90} & \textbf{83.48} \\ \bottomrule
\end{tabular}
}
\caption{Quantitative comparison of video generation models on VBench. \textbf{ImagerySearch} achieves the best average performance across multiple metrics, indicating better alignment and generation quality.}
\label{tb:vbench}
\end{table*}

\begin{figure*}[!ht]
    \centering

    \includegraphics[width=\linewidth]{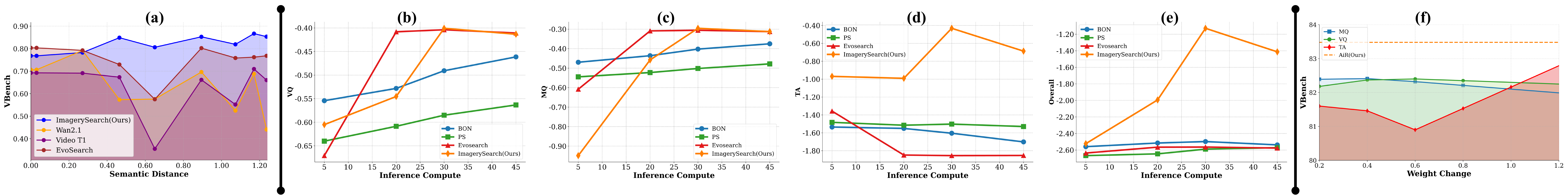}

    \caption{(a) Effect of semantic distance across different models. As semantic distance increases, our method remains the most stable. (b-e) Our AIR consistently delivers superior performance. Scaling behavior of ImagerySeach and baselines as inference-time computation increases. From left to right, the $y$-axes represent the score changes for $\mathrm{MQ}$, $\mathrm{TA}$, $\mathrm{VQ}$, and Overall (VideoAlign~\citep{liu2025videoalign}). (f) Effect of reward weight.}
    \label{fig:all_table}
\end{figure*}

\begin{table*}[!ht]

\centering
\resizebox{\linewidth}{!}{
\begin{tabular}{llccccccc}
\toprule
\multicolumn{2}{c|}{}                                                          & \multicolumn{7}{c}{VBench   (\%)}                                                                                                                                                                                                                                                                                 \\ \cmidrule{3-9} 
\multicolumn{2}{c|}{\multirow{-2}{*}{Model}}                                   & \makecell[c]{Aesthetic\\Quality}                      & \makecell[c]{Background\\Consistency}               & \makecell[c]{Dynamic\\Degree}                         & \makecell[c]{Imaging\\Quality}                        & \makecell[c]{Motion\\Smoothness}                      & \multicolumn{1}{l|}{\makecell[c]{Subject\\consistency}}                    & Average                                \\ \midrule
Baseline                              & \multicolumn{1}{l|}{Wan2.1~\citep{wan2.1}}           & 50.50          & 91.80          & 82.85          & 58.25          & 97.50          & \multicolumn{1}{l|}{90.25}          & 78.53          \\ \midrule
                                      & \multicolumn{1}{l|}{w/o AIR}           & 56.25          & 94.60          & 81.85          & 68.05          & 97.50          & \multicolumn{1}{l|}{94.40}          & 82.11          \\
\multirow{-2}{*}{Modules}              & \multicolumn{1}{l|}{w/o SaDSS}         & 55.35          & 95.10          & 77.20          & 68.00          & 97.60          & \multicolumn{1}{l|}{94.55}          & 81.30          \\ \midrule

                                      & \multicolumn{1}{l|}{0.5}               & 57.25          & 96.15          & 70.00          & 70.75          & 97.45          & \multicolumn{1}{l|}{95.45}          & 81.18          \\
\multirow{-2}{*}{\makecell[l]{SaDSS-static\\weight}} & \multicolumn{1}{l|}{0.9}               & 57.40          & \textbf{96.05}          & 70.00          & 70.80          & 97.55          & \multicolumn{1}{l|}{\textbf{95.50}}          & 81.22          \\ \midrule
                                      & \multicolumn{1}{l|}{BON~\citep{ma2025inference}}               & 57.40          & 95.00          & 83.01          & 68.10          & \textbf{97.70} & \multicolumn{1}{l|}{94.63}          & 82.64          \\
                                      & \multicolumn{1}{l|}{Particle Sampling~\citep{ma2025inference}} & 56.51          & 93.52          & 81.72          & 67.04          & 96.18          & \multicolumn{1}{l|}{93.38}          & 81.39          \\ \cmidrule{2-9} 
\multirow{-3}{*}{Search}              & \textbf{ImagerySearch (Ours)}          & \textbf{57.70} & {96.00} & \textbf{84.05} & \textbf{68.50} & 97.65          & \multicolumn{1}{l|}{{94.70}} & \textbf{83.10} \\ \bottomrule
\end{tabular}
}

\caption{Ablation Study. ``Baseline'' is the plain backbone; ``Modules'' successively add our two novel modules; ``SaDSS-static weight'' denotes the performance obtained when the selection space is kept at a fixed size; ``Search'' swaps in alternative search strategies. The full configuration (ImagerySearch) yields the best performance.}

\label{tb:ab_study}
\end{table*}

\noindent\textbf{Performance on VBench.} For a balanced evaluation, we compare two classes of methods on VBench. The upper rows of Tab.~\ref{tb:vbench} report general generators, while the lower rows list test-time scaling approaches--Video-T1~\citep{liu2025videot1}, EvoSearch~\citep{he2025evosearch}, and our proposed ImagerySearch. All models are evaluated on long-distance prompts from LDT-Bench using the VBench metrics. ImagerySearch achieves the best overall score and ranks highest on the fine-grained \textit{Dynamic Degree}, \textit{Subject Consistency} metrics and so on, indicating its strong ability to preserve prompt fidelity under wide semantic gaps. Fig.~\ref{fig:visual_examples} illustrates this strength: ImagerySearch accurately reproduces both the specified subjects ($e.g.$, \textit{bear}, \textit{controls}) and their associated actions ($e.g.$, \textit{uses}). Additional examples in Appendix D further demonstrate its robustness in handling complex long-distance prompts.

\noindent\textbf{Robustness Analysis Across Semantic Distances.} As illustrated in Fig.~\ref{fig:all_table}(a), our approach maintains nearly constant VBench scores as semantic distance increases, whereas competing methods exhibit pronounced fluctuations. This stability highlights the superior robustness of our model across a wide range of semantic distances. Additional error analysis is provided in the Appendix E.

\subsection{Test-time Scaling Law Analysis} 
We measure the inference-time computation by the number of function evaluations (NFEs). As shown in Fig.~\ref{fig:all_table}(b–d), where performance is assessed with the $\mathrm{MQ}$, $\mathrm{TA}$, and $\mathrm{VQ}$ metrics from VideoAlign \citep{liu2025videoalign}, ImagerySearch exhibits monotonic performance improvements as inference-time computation increases. Notably, on Wan2.1 \citep{wan2.1}, ImagerySearch continues to gain as NFEs grow, whereas baseline methods plateau at roughly $1\times10^{3}$ NFEs (corresponding to the 30th timestep). Computation details are provided in the Appendix F. Moreover, our method shows an even more pronounced advantage in the overall VideoAlign score, as illustrated in Fig.~\ref{fig:all_table}(e).

\subsection{Ablation Study}

\noindent\textbf{Effect of SaDSS and AIR.} As shown in the first three rows of Tab.~\ref{tb:ab_study}, adding either the SaDSS or the AIR module individually already surpasses the baseline, while combining SaDSS with AIR achieves the best performance, confirming the complementary nature of semantic guidance and adaptive selection.

\noindent\textbf{Effect of Search Space Size.} The \textit{SaDSS–static weight} rows in Tab.~\ref{tb:ab_study} compare fixed and dynamic search-space configurations. With static weights of 0.5, and 0.9, performance improves gradually, reaching a VBench score of 81.22\%. In contrast, the dynamic approach attains a markedly higher score of 83.48\%, demonstrating its superior ability to optimize the search space and thus boost model performance.

\noindent\textbf{Effect of Search Strategy.} The \textit{Search} rows in Tab.~\ref{tb:ab_study} compare different search strategies ($e.g.$, BON, Particle Sampling~\citep{ma2025inference}). The experimental results demonstrate that our search strategy delivers the best performance.

\noindent\textbf{Effect of Reward Dynamic Adjustment Mechanism.} Fig.~\ref{fig:all_table}(f) demonstrates the impact of varying reward weights on VBench scores across different models ($\mathrm{MQ}$, $\mathrm{TA}$, $\mathrm{VQ}$). As weights change from 0.2 to 1.2, $\mathrm{TA}$ shows notable improvement while $\mathrm{MQ}$ and $\mathrm{VQ}$ maintain relatively stable performance. The consistent superiority of the Ours approach, represented by the dashed line, underscores the effectiveness of dynamic reward adjustment, achieving optimal performance irrespective of weight changes.

\section{Conclusion}
\label{sec:conclusion}

In this study, we propose ImagerySearch, an adaptive test-time search method that improves video-generation quality for long-distance semantic prompts drawn from imaginative scenarios. Additionally, we present LDT-Bench, the first benchmark designed to evaluate such challenging prompts. ImagerySearch attains state-of-the-art results on both VBench and LDT-Bench, with especially strong gains on LDT-Bench, demonstrating its effectiveness for text-to-video generation under long-range semantic conditions. In future, we will explore more flexible reward mechanisms to further enhance video-generation performance.

\bibliographystyle{plainnat}
\bibliography{iclr2026_conference}

\medskip

\clearpage
\setcounter{page}{1}
\renewcommand\thefigure{S\arabic{figure}}
\setcounter{figure}{0}
\renewcommand\thetable{S\arabic{table}}
\setcounter{table}{0}

\appendix                            
\renewcommand{\thesection}{\Alph{section}}
\setcounter{section}{0}

\section{The Selection of Imagery Schedule} 
\label{app:imageryschedule}
\begin{figure*}[t!]
    \centering
    \includegraphics[width=1\linewidth]{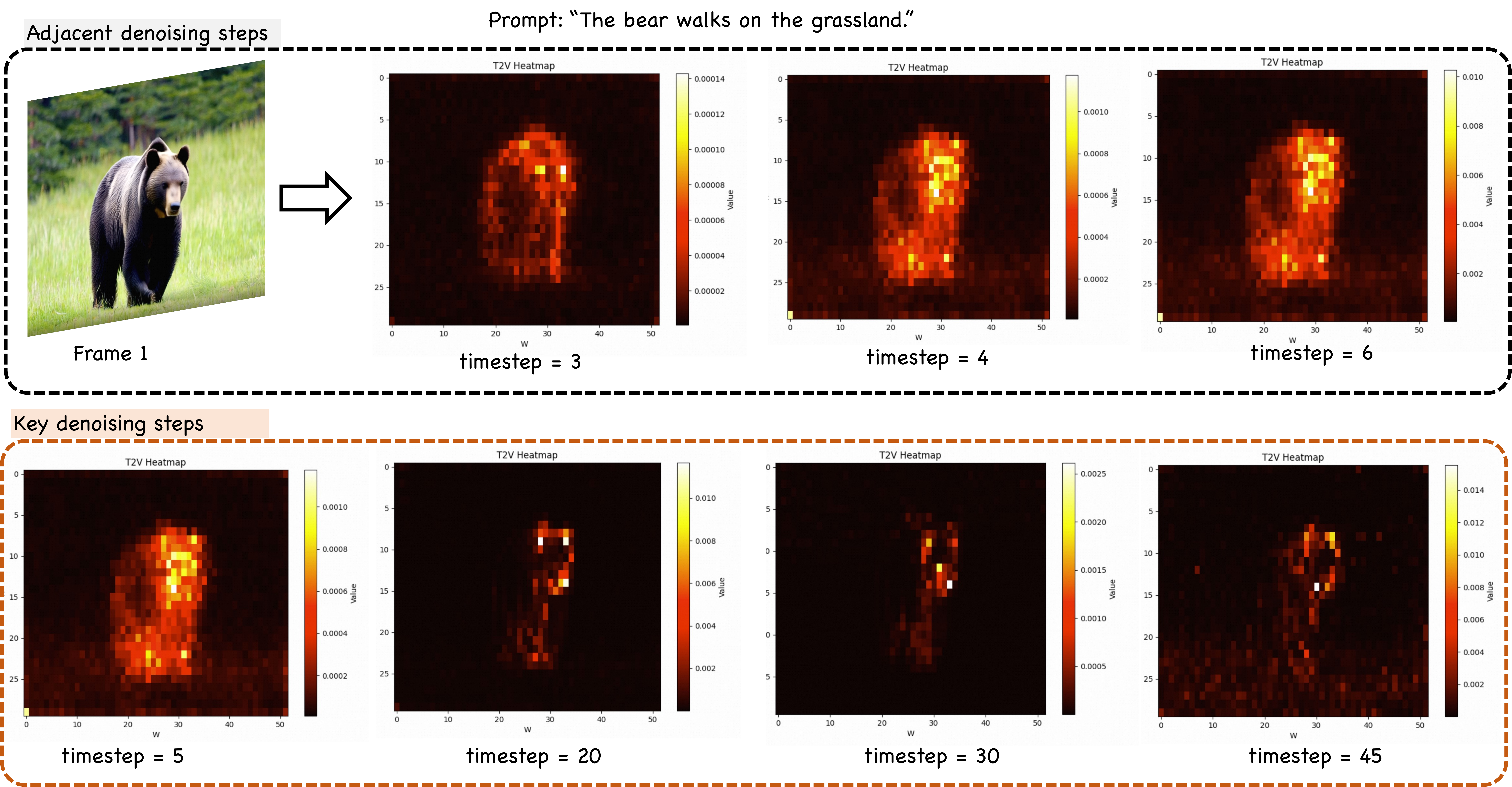}
    \caption{Imagery schedule. The heatmaps visualize 13th-layer attention projected onto the first video frame at successive denoising steps. Adjacent steps show nearly identical focus regions, whereas only a few key steps exhibit pronounced changes. Concentrating analysis and search on these pivotal steps therefore captures the prompt-to-frame semantic correspondence more efficiently.}
    \label{fig:imagery_schedule}
\end{figure*}

As illustrated in Fig.~\ref{fig:imagery_schedule}, we observe that adjacent denoising steps modify the latent video only marginally; substantial deviations from earlier stages emerge only at several pivotal steps. To improve generation efficiency, we therefore trigger \textit{ImagerySearch} at a limited set of noise levels, $\mathcal{S}=\{5,\,20,\,30,\,45\},$ which we term the \emph{Imagery Schedule}. This schedule specifies the exact timesteps at which \textit{ImagerySearch} is invoked.

\section{More Details About Imagery Evaluation Metrics}   
\label{app:metrics}
\subsection{More Text Encoders.} In our current implementation, T5 serves three purposes: it encodes the key entities in each prompt, measures their semantic distances, and then uses those distances to adjust the search space and reward weights during generation. The same pipeline can be run with a CLIP text encoder~\citep{radford2021clip, blattmann2023stablevideodiffusionscaling}. Trained on large-scale image–text pairs, CLIP yields text embeddings whose cosine distances correlate well with visual concepts, so these distances can play exactly the same role in deciding when to expand or shrink the search space. In addition, CLIP similarities are widely used as a measure of text–image or text–video alignment, which makes them a natural choice for the alignment term in our reward function~\citep{stam2023stable}. Because CLIP, like T5, produces a fixed-length vector in a single forward pass, it can be swapped in as a drop-in replacement without changing any downstream components while fully preserving the effectiveness of our adaptive search and reward mechanisms.

\subsection{More Analysis about Prompt Suite. }
\begin{figure*}[t!]
    \centering
    \includegraphics[width=1\linewidth]{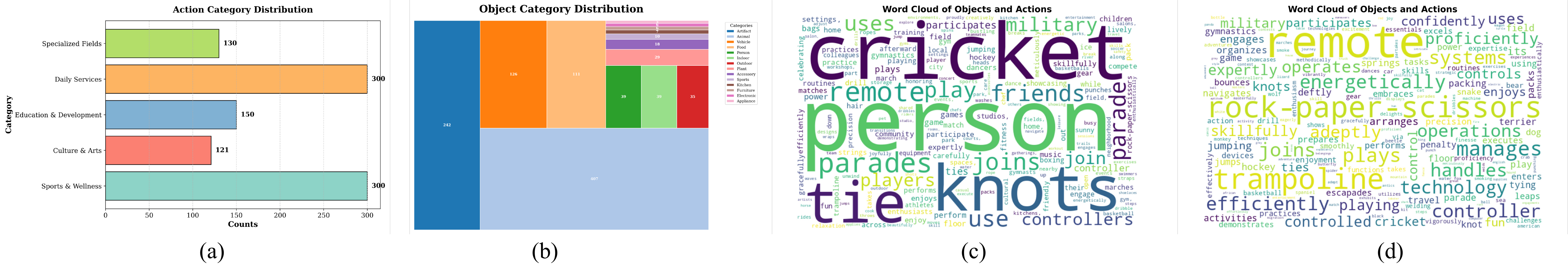}
    \caption{LDT-Bench prompt suite analysis: (a) Action super-category distribution shown as a horizontal bar chart. (b) Object super-category distribution displayed as a treemap, with area proportional to class count. (c) Word cloud highlighting the most frequent object-action prompts. (d) Word cloud highlighting the most frequent action-action prompts.}
    \label{fig:prompt_analysis}
\end{figure*}

As shown in Fig.~\ref{fig:prompt_analysis}, we provide a multi-faceted overview of the \emph{LDT-Bench} prompt suite and underscore its advantages for long-distance semantic evaluation.  
\textbf{(a)}~Examining the distribution of actions, a pronounced long-tail pattern emerges: of the five super-categories, \emph{Sports \& Wellness} and \emph{Daily Services} each supply 300 prompts, ensuring ample coverage of everyday yet highly diverse actions.  
\textbf{(b)}~For objects, a treemap of 14 super-categories—scaled by instance count—reveals that \emph{Animal} and \emph{Artifact} jointly exceed half of all samples, while still leaving room for rarer classes; this balance of head and tail categories is largely missing in prior benchmarks.  
\textbf{(c)}~The object word cloud (after stop-word filtering) highlights high-frequency nouns such as \emph{cricket}, \emph{person}, and \emph{remote}, evidencing fine-grained lexical diversity across domains.  
\textbf{(d)}~The action word cloud reveals a wide semantic span—verbs like \emph{play}, \emph{join}, \emph{use}, and \emph{handle}—that challenges models to cope with imaginative, long-distance dependencies.  

Taken together, these statistics show that \emph{LDT-Bench} not only covers a richer mix of objects and actions than existing datasets but also accentuates long-distance semantic relationships that current models find most difficult, making it a uniquely effective testbed for stress-testing creative video generation systems.

\subsection{ImageryQA Implementation Details.} 
\begin{figure*}[t!]
    \centering
    \includegraphics[width=0.9\linewidth]{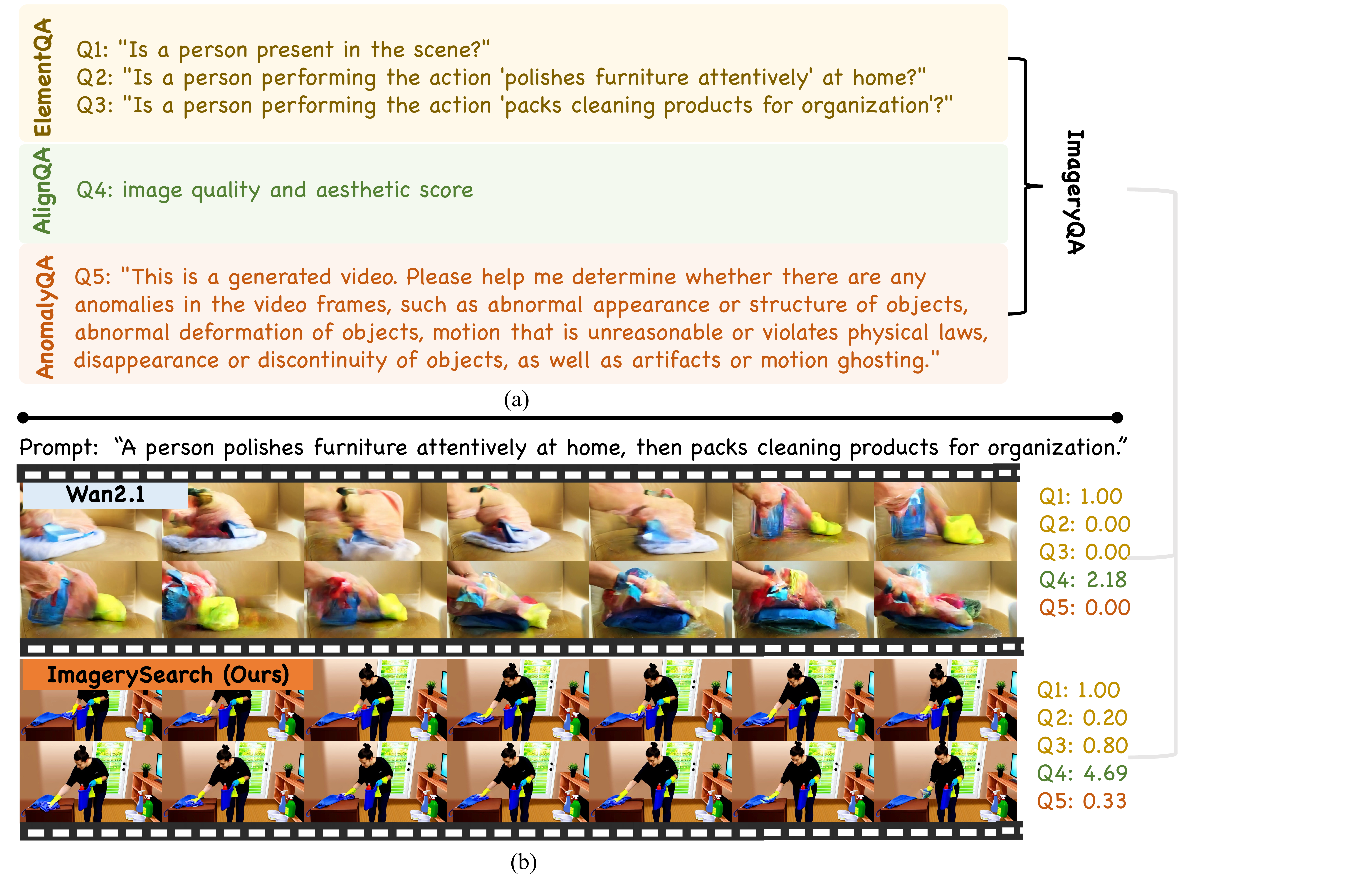}
    \caption{Evaluation with \emph{ImageryQA}. \textbf{(a)}~We design a structured question set \emph{ImageryQA}, consisting of \textit{ElementQA}, \textit{AlignQA}, and \textit{AnomalyQA}.\textbf{(b)}~Comparison between Wan2.1 and ImagerySearch on the same prompt. Wan2.1 fails to depict a person and the actions described, resulting in low aesthetic quality (Q4) and visual anomalies (Q5). In contrast, ImagerySearch successfully captures both actions--\textit{polishing furniture} and \textit{packing cleaning products}--scoring higher in both Q4 and Q5.}
    \label{fig:imageryQA}
\end{figure*}

As described in Sec. 4.2 of the paper, our metric is primarily composed of three components: ElementQA, AlignQA, and AnomalyQA (Fig.~\ref{fig:imageryQA} (a)). In this subsection, we provide further clarification using specific examples and illustrating the metric computation process.

As shown in Fig.~\ref{fig:imageryQA} (b), given the evaluation prompt, ``A person polishes furniture attentively at home, then packs cleaning products for organization.'', two videos generated by different video generation models. First, ElementQA formulates questions based on the objects and actions within the prompt, $i.e.$, ``person,'' ``polishes furniture,'' and ``packs cleaning products for organization'', resulting in the questions Q1, Q2, and Q3 in Fig.~\ref{fig:imageryQA}.
Next, AlignQA assesses the first frame of each video in terms of image quality and aesthetics.
Finally, AnomalyQA evaluates abnormal events in both videos, as illustrated by Q5 in Fig.~\ref{fig:imageryQA}.

Based on these questions, we employ different MLLMs and answer strategies. Recent studies~\citep{feng2025narrlv, liu2025videoalign, wu2024qalign, zheng2025vbench,wu2024finger} suggest that for questions with inherent uncertainty, having a general-purpose MLLM~\citep{qwen2.5, openai2024gpt4o} answer the same question multiple times and averaging the results yields more reliable evaluations. Therefore, for ElementQA, we prompt Qwen2.5-VL-72B-Instruct~\citep{qwen2.5} to answer each question five times. 
For AnomalyQA, considering the higher cost of GPT-4o~\citep{openai2024gpt4o}, we collect three responses per question. For Q-Align~\citep{wu2023q} in AlignQA, since it is a dedicated model trained for aesthetic quality assessment and directly outputs a quantitative score, we use a single response.

\begin{figure*}
    \centering
    \includegraphics[width=0.9\linewidth]{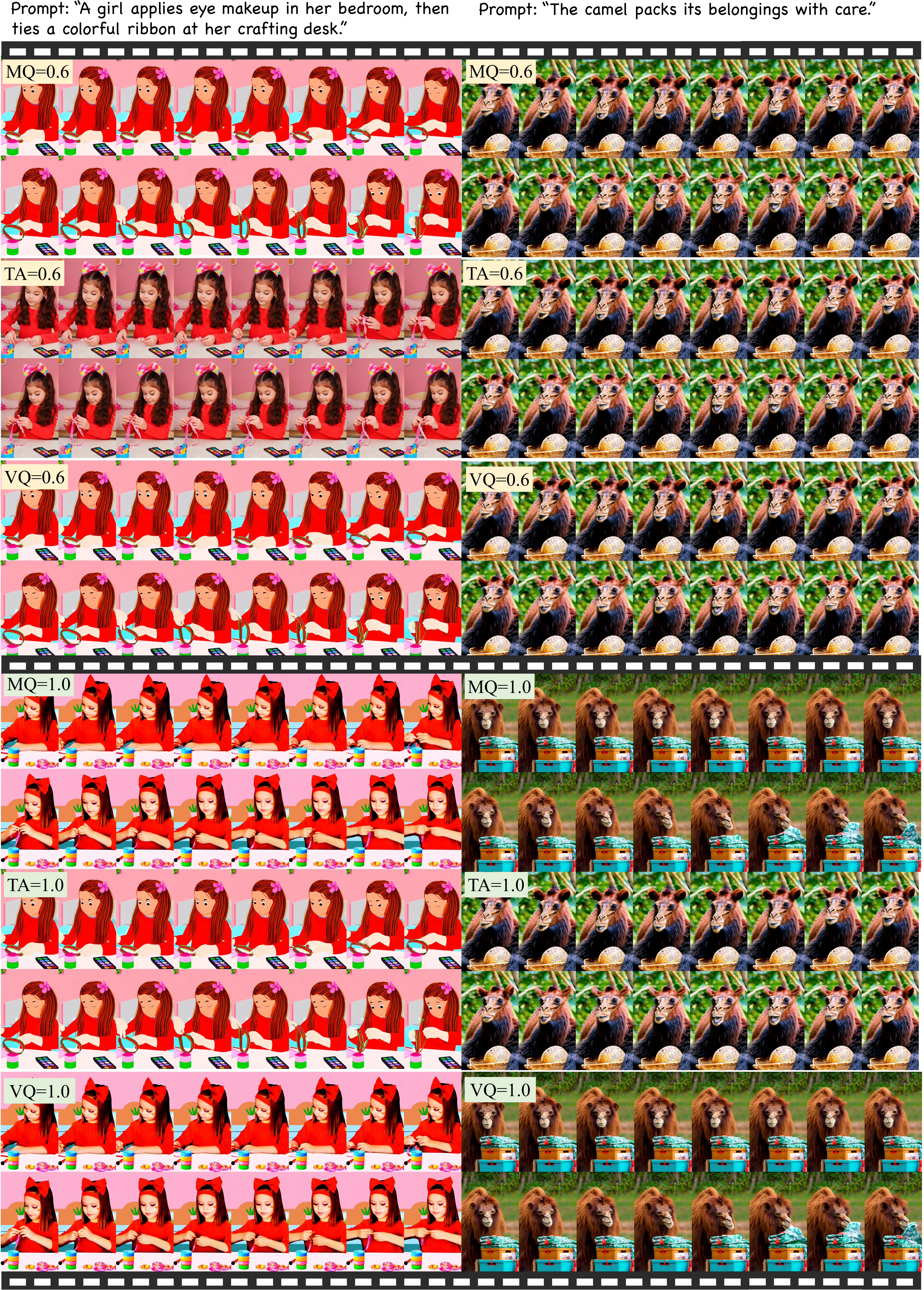}
    \caption{Reward-Weight Analysis. The left of figure shows an \textbf{action–action} example and the right of figure is an \textbf{object–action} one, visualizing the videos under different weight settings. $MQ$ and $VQ$ follow almost identical trends, whereas $TA$ moves in the opposite direction. Accordingly, we fix the $MQ$ and $VQ$ coefficients to 1 and vary the $TA$ coefficient with the prompt, selecting videos that better fit imaginative scenarios.}
    \label{fig:reward_weight}
\end{figure*}

\section{Experimental Setup--Model details} 
\textbf{Parameter settings.} In our implementation, the baseline model is Wan2.1-1.3B~\citep{wan2.1}. And we set the imagery schedule to $\{5, 20, 30, 45\}$ and set the imagery size schedule to $\{10, 5, 5, 5, 5\}$. As shown in Fig.~\ref{fig:reward_weight}, $VQ$ and $MQ$ exhibit the same selection trend, whereas $TA$ shows the opposite. Therefore, regarding the parameters in Equation (5), we set $\beta=\gamma=1.0$, and $\alpha$ are dynamically adjusted.

\section{More Examples}

\begin{figure*}
    \centering
    \includegraphics[width=0.9\linewidth]{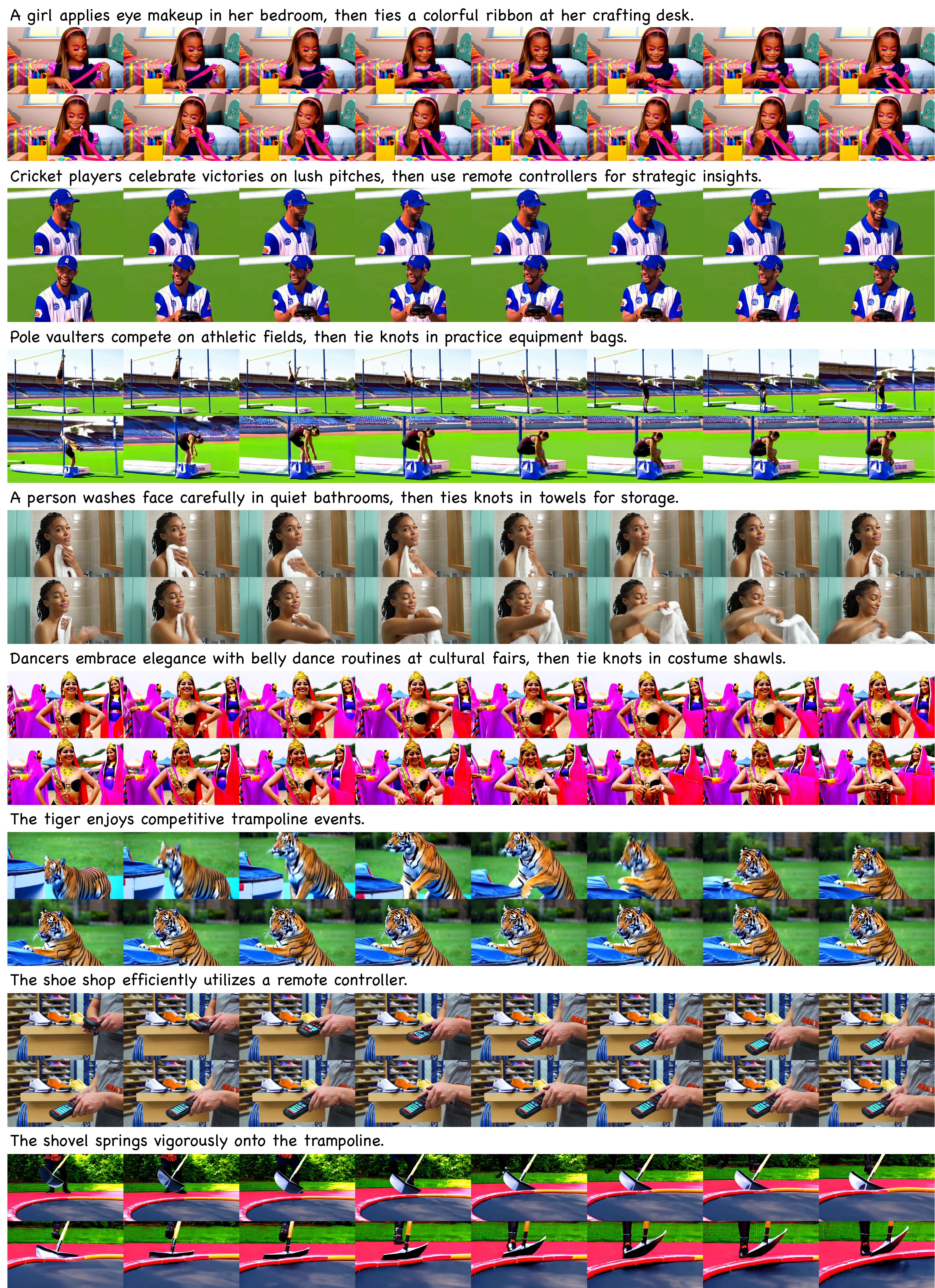}
    \caption{More examples on LDT-Bench. The images below the prompt show the result of frame sampling, where 16 frames are uniformly extracted from a 33-frame video.}
    \label{fig:more_examples}
\end{figure*}

\begin{figure*}
    \centering
    \includegraphics[width=0.9\linewidth]{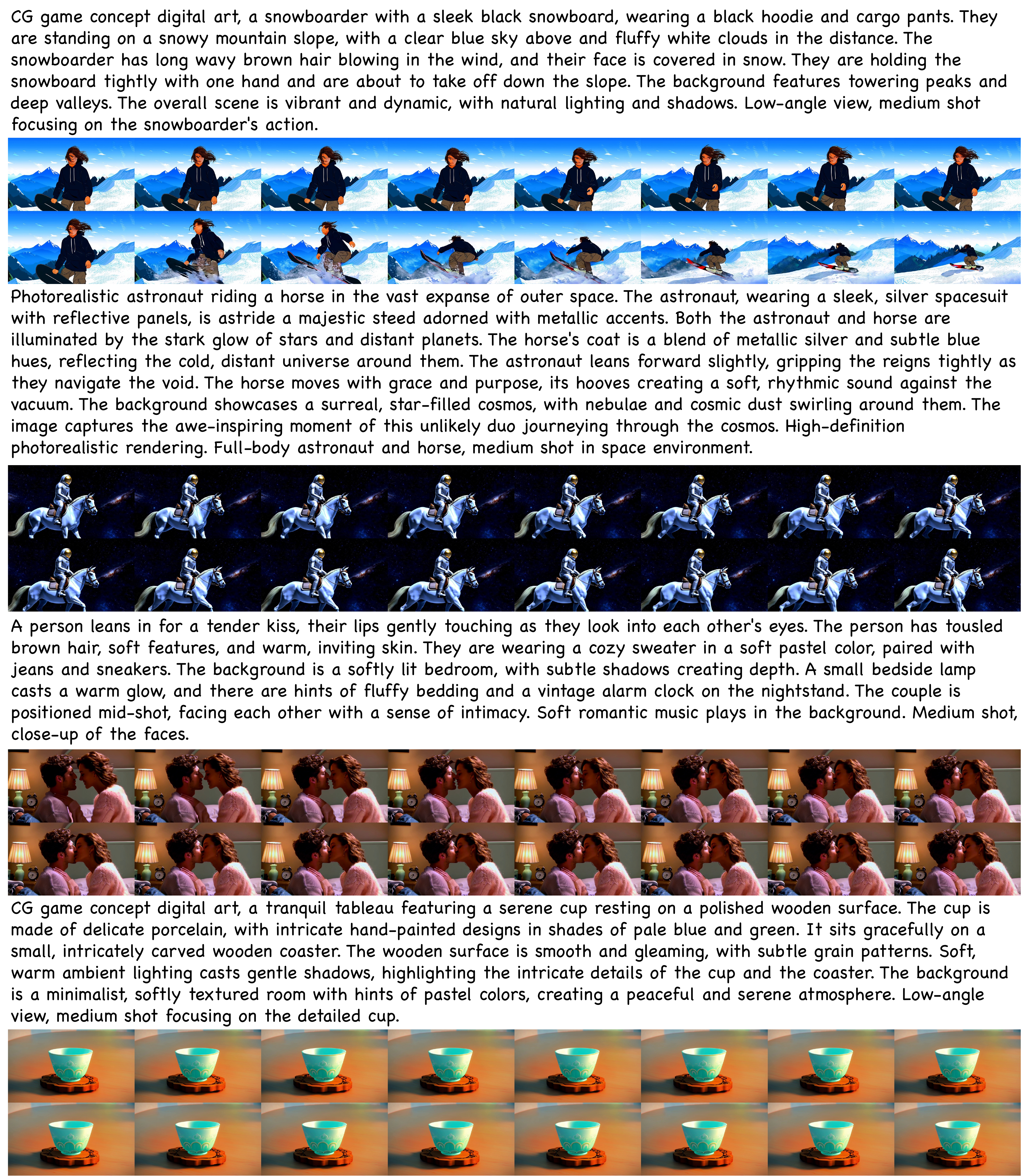}
    \caption{More examples on VBench (Part I). The images below the prompt show the result of frame sampling, where 16 frames are uniformly extracted from a 33-frame video.}
    \label{fig:more_examples_2}
\end{figure*}

\begin{figure*}
    \centering
    \includegraphics[width=0.9\linewidth]{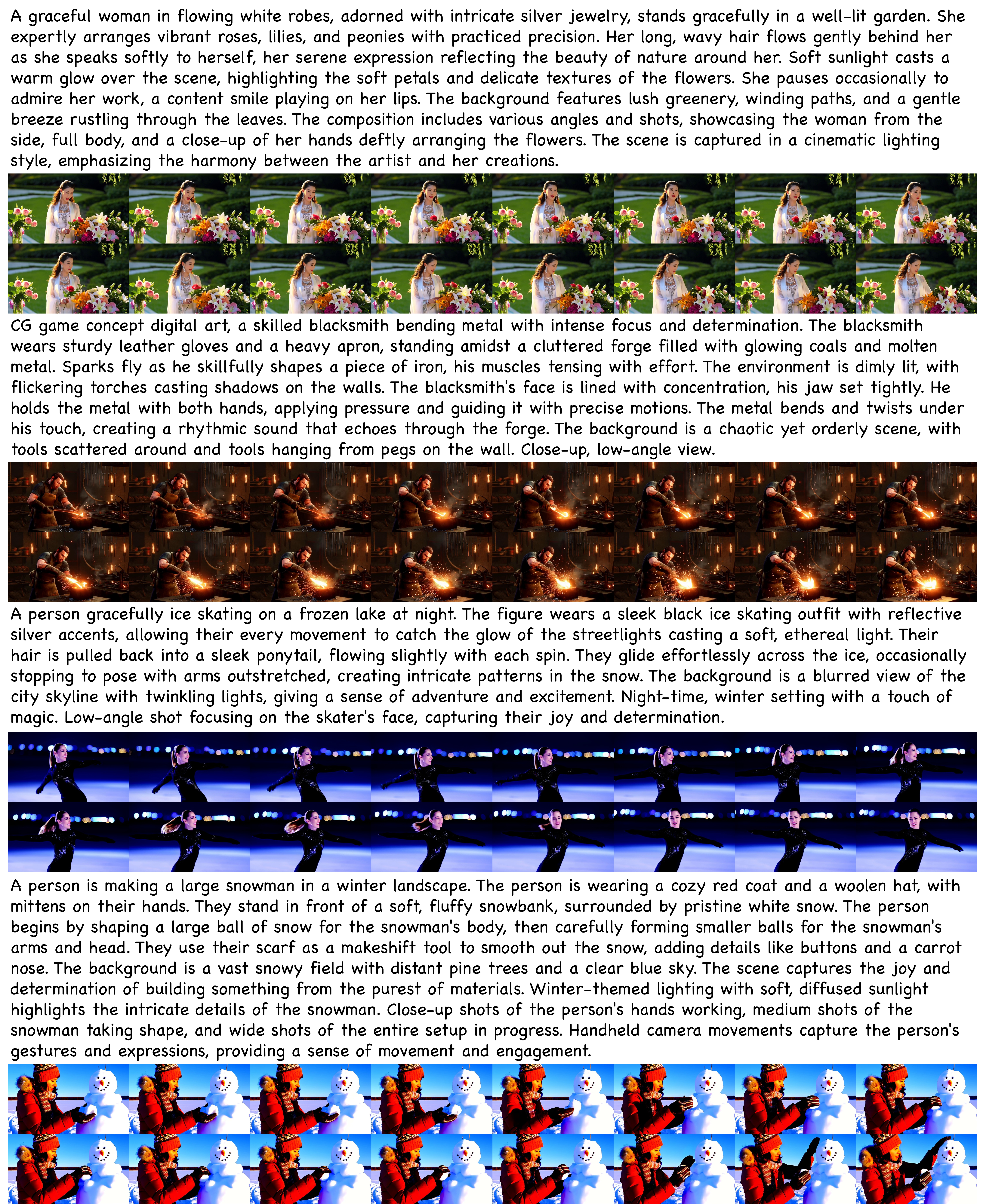}
    \caption{More examples on VBench (Part II). The images below the prompt show the result of frame sampling, where 16 frames are uniformly extracted from a 33-frame video.}
    \label{fig:more_examples_3}
\end{figure*}

Additional qualitative examples are provided in Fig.~\ref{fig:more_examples}, Fig.~\ref{fig:more_examples_2}, and
Fig.~\ref{fig:more_examples_3}. Specifically, Fig.~\ref{fig:more_examples} reports results on \textsc{LDT-Bench}, where the first five rows correspond to \emph{action–action} prompts and the last three to \emph{object–action} prompts. Fig.~\ref{fig:more_examples_2} and Fig.~\ref{fig:more_examples_3} show further \emph{action–action} cases drawn from VBench. Across all examples, our method produces vivid and coherent videos, even under long-distance semantic prompts, illustrating its capacity to handle challenging imaginative scenarios.

\section{Error Analysis}
\begin{figure}
    \centering
    \includegraphics[width=0.7\linewidth]{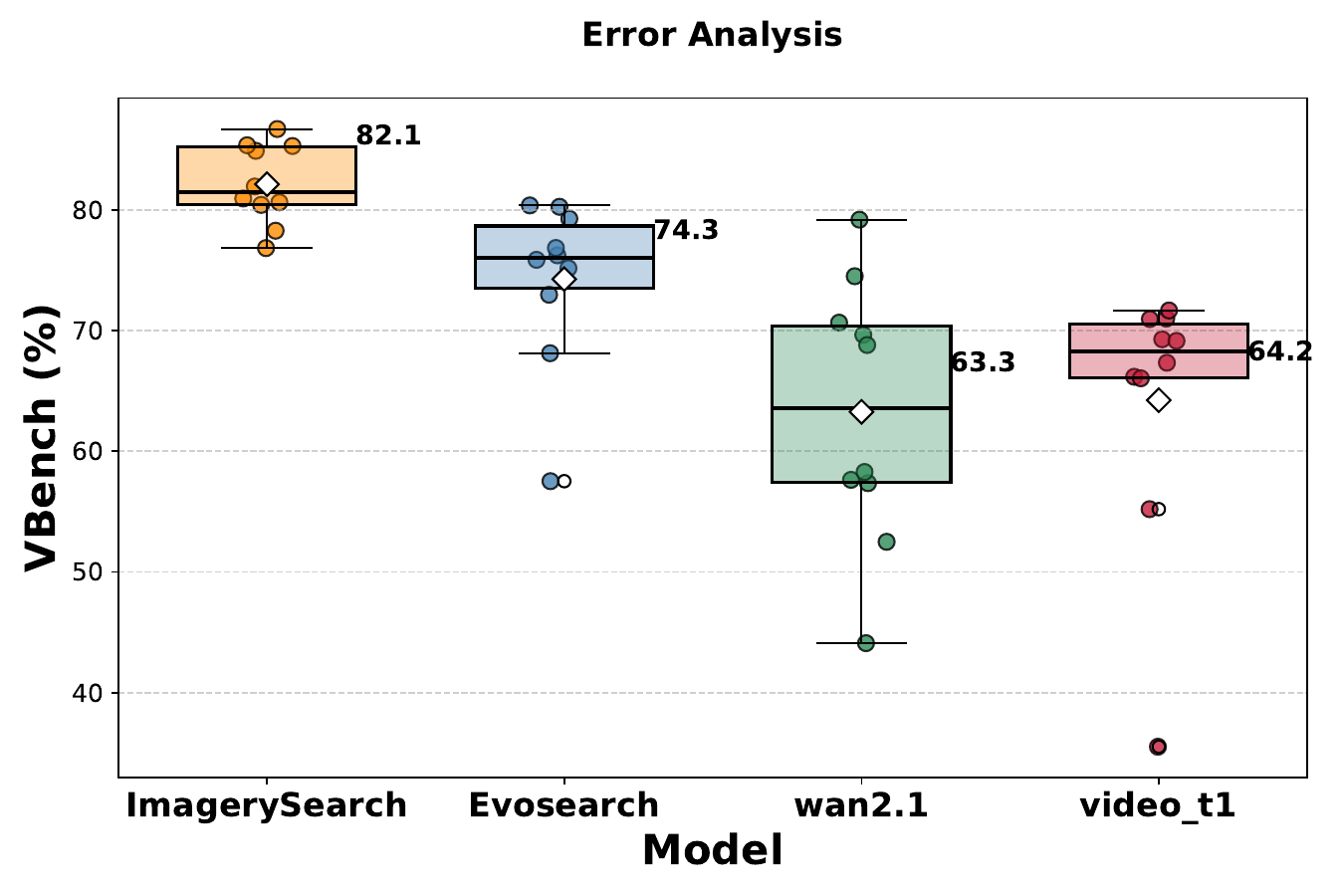}
    \caption{Error analysis about VBench scores on long-distance semantic prompts. Each box shows the score distribution for one model (mean marked by a white diamond); individual data points are overlaid in matching colors. ImagerySearch (orange) attains the highest mean with the tightest spread, while the other methods exhibit lower central tendencies and larger variances.}
    \label{fig:error_analysis}
\end{figure}

In the VBench~\citep{huang2024vbench} error analysis (Fig.~\ref{fig:error_analysis}), ImagerySearch shows a higher mean score with a tighter interquartile range, indicating more stable performance across prompts. Evosearch~\citep{he2025evosearch} attains a comparable median but displays greater dispersion, whereas wan2.1~\citep{wan2.1} and Video-T1~\citep{liu2025videot1} exhibit lower central scores and wider quartile spans. Overall, dynamically adjusting the search space and rewarding by semantic distance helps maintain generation quality while reducing sensitivity to prompt difficulty.
\clearpage
\bigskip
\noindent 

\end{document}